\documentclass[11pt,a4paper]{article}
\usepackage[hyperref]{emnlp2018}
\usepackage{times}
\usepackage{latexsym}
\usepackage{url}

\usepackage{graphicx}
\usepackage{amsmath}
\usepackage[boldmath]{numprint}
\usepackage{caption}
\usepackage{xcolor}

\usepackage{enumitem}

\usepackage[titletoc,title]{appendix}

\aclfinalcopy 

\title{Learning Personas from Dialogue with\\
Attentive Memory Networks}

\author{Eric Chu \thanks{The first two authors contributed equally to this work.}  \\ 
  MIT Media Lab \\
  {\tt echu@mit.edu} \\\And
  Prashanth Vijayaraghavan \footnotemark[1] \\
    MIT Media Lab \\
  {\tt pralav@mit.edu}\\\And
  Deb Roy \\
    MIT Media Lab \\
  {\tt dkroy@media.mit.edu} \\}

\date{}

\begin{document}

\maketitle
\begin{abstract}
The ability to infer persona from dialogue can have applications in areas ranging from computational narrative analysis to personalized dialogue generation. We introduce neural models to learn persona embeddings in a supervised character trope classification task. The models encode dialogue snippets from IMDB into representations that can capture the various categories of film characters.  The best-performing models use a multi-level attention mechanism over a set of utterances. We also utilize prior knowledge in the form of textual descriptions of the different tropes. We apply the learned embeddings to find similar characters across different movies, and cluster movies according to the distribution of the embeddings. The use of short conversational text as input, and the ability to learn from prior knowledge using memory, suggests these methods could be applied to other domains.
\end{abstract}

\section{Introduction}
Individual personality plays a deep and pervasive role in shaping social life. Research indicates that it can relate to the professional and personal relationships we develop \cite{barrick1993autonomy}, \cite{shaver1992attachment}, the technological interfaces we prefer \cite{nass2000does}, the behavior we exhibit on social media networks \cite{selfhout2010emerging}, and the political stances we take \cite{jost2009personality}.

With increasing advances in human-machine dialogue systems, and widespread use of social media in which people express themselves via short text messages, there is growing interest in systems that have an ability to understand different personality types. Automated personality analysis based on short text analysis could open up a range of potential applications, such as dialogue agents that sense personality in order to generate more interesting and varied conversations.

We define persona as a person's social role, which can be categorized according to their conversations, beliefs, and actions. To learn personas, we start with the character tropes data provided in the CMU Movie Summary Corpus by \cite{bamman2014learning}. It consists of 72 manually identified commonly occurring character archetypes and examples of each. In the character trope classification task, we predict the character trope based on a batch of dialogue snippets.

In their original work, the authors use Wikipedia plot summaries to learn latent variable models that provide a clustering from words to topics and topics to personas -- their persona clusterings were then evaluated by measuring similarity to the ground-truth character trope clusters. We asked the question -- could personas also be inferred through dialogue? Because we use quotes as a primary input and not plot summaries, we believe our model is extensible to areas such as dialogue generation and conversational analysis.

\begin{table*}[ht!]
\small
\begin{center}
\begin{tabular}{|c|c|c|}
\hline \bf Character Trope & \bf Character   & \bf Movie \\ \hline
Corrupt corporate executive & Les Grossman & Tropic Thunder  \\
Retired outlaw & Butch Cassidy & Butch Cassidy and the Sundance Kid \\
Lovable rogue & Wolverine & X-Men \\
\hline
\end{tabular}
\end{center}
\caption{\label{trope_data}Example tropes and characters}
\normalsize
\end{table*}

Our contributions are:
\begin{enumerate}[topsep=0pt]
  \setlength{\itemsep}{1pt}
  \setlength{\parskip}{0pt}
  \setlength{\parsep}{0pt}
  \item Data collection of IMDB quotes and character trope descriptions for characters from the CMU Movie Summary Corpus.
  \item Models that greatly outperform the baseline model in the character trope classification task. Our experiments show the importance of multi-level attention over words in dialogue, and over a set of dialogue snippets. 
  \item We also examine how prior knowledge in the form of textual descriptions of the persona categories may be used. We find that a `Knowledge-Store' memory initialized with descriptions of the tropes is particularly useful. This ability may allow these models to be used more flexibly in new domains and with different persona categories.
\end{enumerate}

\section{Related Work}
Prior to data-driven approaches, personalities were largely measured by asking people questions and assigning traits according to some fixed set of dimensions, such as the Big Five traits of openness, conscientiousness, extraversion, agreeability, and neuroticism \cite{tupes1992recurrent}. Computational approaches have since advanced to infer these personalities based on observable behaviors such as the actions people take and the language they use \cite{golbeck2011predicting}.

Our work builds on recent advances in neural networks that have been used for natural language processing tasks such as reading comprehension \cite{sukhbaatar2015end} and dialogue modeling and generation \cite{vinyals2015neural, li2016persona, shang2015neural}. This includes the growing literature in attention mechanisms and memory networks \cite{bahdanau2014neural, sukhbaatar2015end, kumar2016ask}.

The ability to infer and model personality has applications in storytelling agents, dialogue systems, and psychometric analysis. In particular, personality-infused agents can help ``chit-chat'' bots avoid repetitive and uninteresting utterances \cite{walker1997improvising, mairesse2007personage, li2016persona, zhang2018personalizing}. The more recent neural models do so by conditioning on a `persona' embedding -- our model could help produce those embeddings.

Finally, in the field of literary analysis, graphical models have been proposed for learning character personas in novels \cite{flekova2015personality, srivastava2016inferring}, folktales \cite{valls2014toward}, and movies \cite{bamman2014learning}. However, these models often use more structured inputs than dialogue to learn personas. 

%
%
\section{Datasets}
Characters in movies can often be categorized into archetypal roles and personalities. To understand the relationship between dialogue and personas, we utilized three different datasets for our models: (a) the Movie Character Trope dataset, (b) the IMDB Dialogue Dataset, and (c) the Character Trope Description Dataset. We collected the IMDB Dialogue and Trope Description datasets, and these datasets are made publicly available \footnote{\url{https://pralav.github.io/emnlp\_personas/}}.

\subsection{Character Tropes Dataset}
\label{char_trope}
The CMU Movie Summary dataset provides tropes commonly occurring in stories and media \cite{bamman2014learning}.
There are a total of 72 tropes, which span 433 characters and 384 movies. Each trope contains between 1 and 25 characters, with a median of 6 characters per trope. Tropes and canonical examples are shown in Table \ref{trope_data}. 

\subsection{IMDB Dialogue Snippet Dataset}
\label{imdb_diag}
\begin{figure}
\includegraphics[width=\columnwidth]{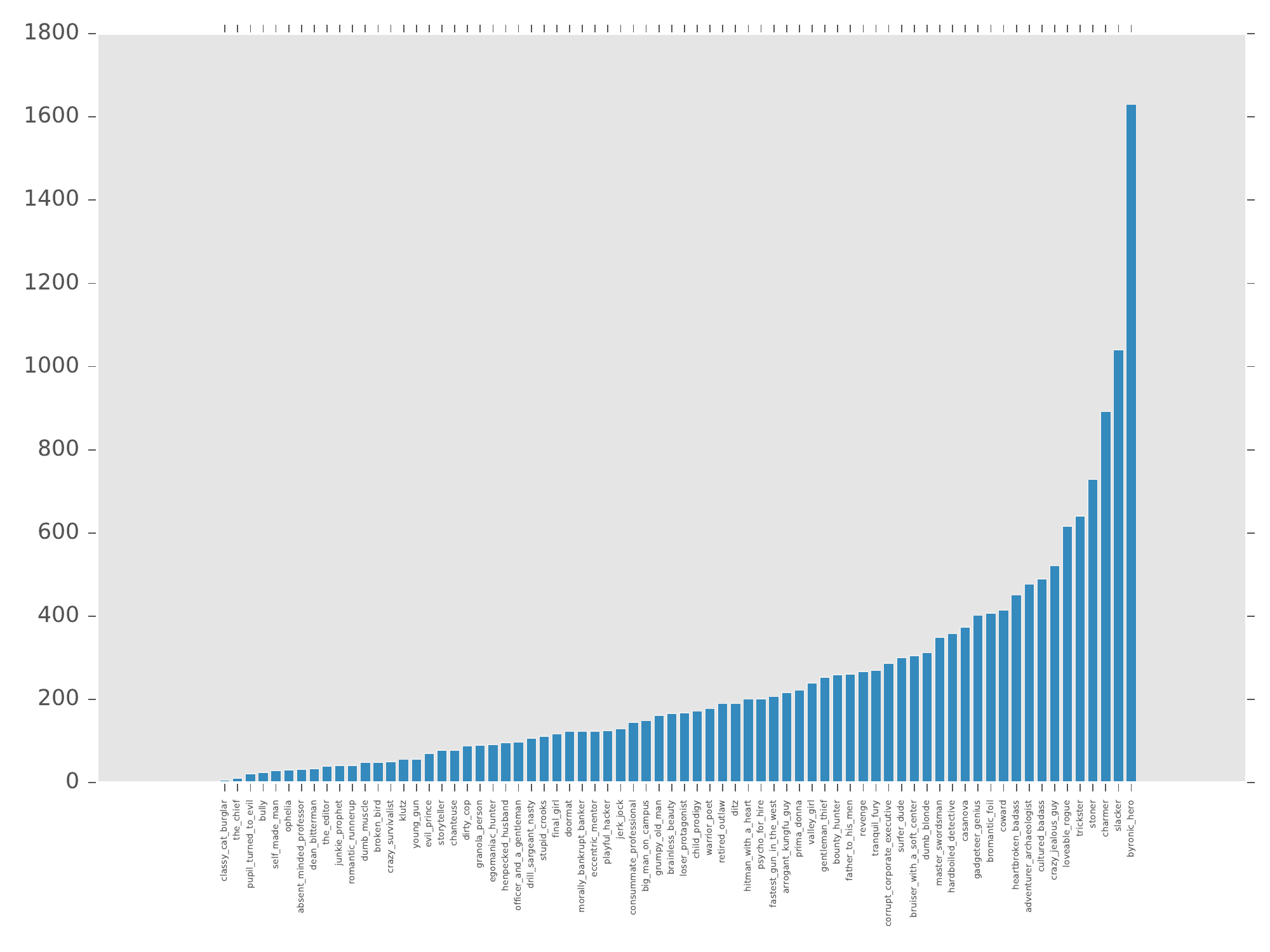}
\caption{\label{trope_quotes_dist}Number of IMDB dialogue snippets per trope}
\end{figure}

To obtain the utterances spoken by the characters, we crawled the IMDB Quotes page for each movie. Though not every single utterance spoken by the character may be available, as the quotes are submitted by IMDB users, many quotes from most of the characters are typically found, especially for the famous characters found in the Character Tropes dataset. The distribution of quotes per trope is displayed in Figure \ref{trope_quotes_dist}. Our models were trained on 13,874 quotes and validated and tested on a set of 1,734 quotes each. 

\begin{figure}
\includegraphics[width=\linewidth]{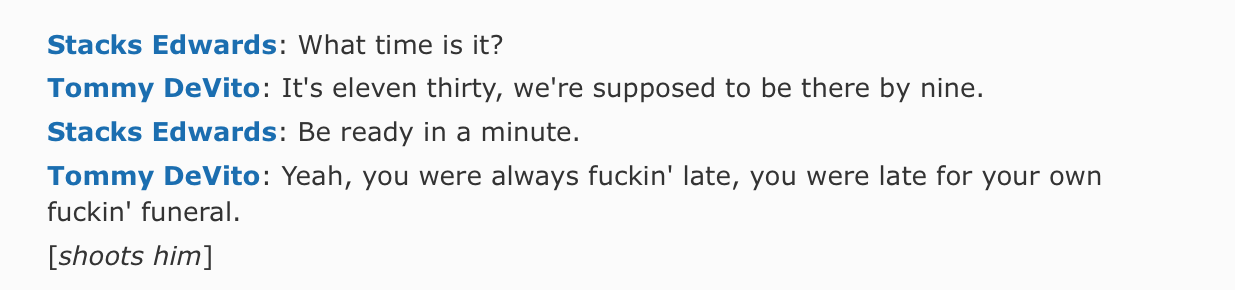}
\caption{\label{dialog_snippet}Example IMDB dialogue snippet containing multiple characters and context.}
\end{figure}

We refer to each IMDB quote as a (contextualized) dialogue snippet, as each quote can contain several lines between multiple characters, as well as italicized text giving context to what might be happening when the quote took place. Figure \ref{dialog_snippet} show a typical dialogue snippet. 70.3\% of the quotes are multi-turn exchanges, with a mean of 3.34 turns per multi-turn exchange.  While the character's own lines alone can be highly indicative of the trope, our models show that accounting for context and the other characters' lines and context improves performance. The context, for instance, can give clues to typical scenes and actions that are associated with certain tropes, while the other characters' lines give further detail into the relationship between the character and his or her environment.

\subsection{Character Trope Description Dataset}
\label{trope_desc}
We also incorporate descriptions of each of the character tropes by using the corresponding descriptions scraped from TVTropes\footnote{\url{http://tvtropes.org}}. Each description contains several paragraphs describing typical characteristics, actions, personalities, etc. about the trope. As we demonstrate in our experiments, the use of these descriptions improves classification performance. This could allow our model to be applied more flexibly beyond the movie character tropes -- as one example, we could store descriptions of personalities based on the Big Five traits in our Knowledge-Store memory.

%
%
\section{Problem Formulation}
\label{prob}
Our goal is to train a model that can take a batch of dialogue snippets from the IMDB dataset and predict the character trope.

Formally, let $N_P$ be the total number of character tropes in the character tropes dataset. Each character $C$ is associated with a corresponding ground-truth trope category $P$. Let $S=(D, E, O)$ be a dialog snippet associated with a character $C$, where $D=[w_{D_1}, w_{D_2}...,w_{D_T}]$ refers to the character's own lines,  $E=[w_{E_1}, w_{E_2}...,w_{E_T}]$ is the contextual information and $O=[w_{O_1}, w_{O_2}...,w_{O_T}]$ denotes the other characters' lines. We define all three components of $S$ to have fixed sequence length $T$ and pad when necessary. Let $N_S$ be the total number of dialogue snippets for a trope. We sample a set of $N_{diag}$  (where $N_{diag} \ll N_S$) snippets from $N_S$ snippets related to the trope as inputs to our model.

%
%
\section{Attentive Memory Network}

\begin{figure*}[ht!]
\includegraphics[width=\linewidth]{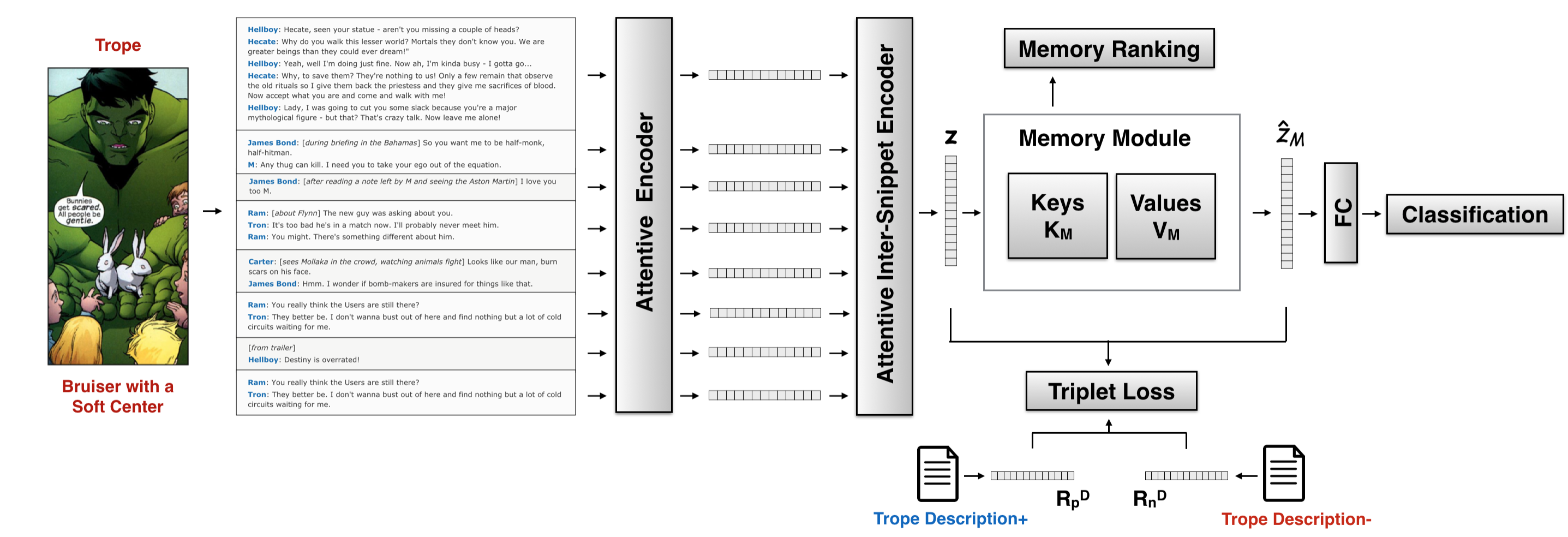}
\caption{\label{nanm}Illustration of the Attentive Memory Network. The network takes dialogue snippets as input and predicts its associated character trope. In this example, dialogue snippets associated with the character trope ``Bruiser with a Soft Corner'' is given as input to the model.}
\end{figure*}

The Attentive Memory Network consists of two major components: (a) Attentive Encoders, and (b) a Knowledge-Store Memory Module.  Figure \ref{nanm} outlines the overall model. We describe the components in the following sections.

\subsection{Attentive Encoders}
Not every piece of dialogue may be reflective of a latent persona. In order to learn to ignore words and dialogue snippets that are not informative about the trope we use a multi-level attentive encoder that operates at (a) the individual snippet level, and (b) across multiple snippets. 

\subsubsection*{Attentive Snippet Encoder}\label{snippet_enc}
The snippet encoder extracts features from a single dialogue snippet $S$, with attention over the words in the snippet. A snippet $S=(D,E,O)$ is fed to the encoder to extract features from each of these textual inputs and encode them into an embedding space. We use a recurrent neural network as our encoder, explained in detail in Section \ref{encoder}. In order to capture the trope-reflective words from the input text, we augment our model with a self-attention layer which scores each word in the given text for its relevance. Section \ref{attention} explains how the attention weights are computed. The output of this encoder is an encoded snippet embedding $S^e=(D^e, E^e, O^e)$.

\subsubsection*{Attentive Inter-Snippet Encoder}\label{isnippet_enc}
As shown in Figure \ref{nanm}, the $N_{diag}$ snippet embeddings $S^e$ from the snippet encoder are fed to our inter-snippet encoder. This encoder captures inter-snippet relationship using recurrence over the $N_{diag}$ snippet embeddings for a given trope and determines their importance. Some of the dialogue snippets may not be informative about the trope, and the model learns to assign low attention scores to such snippets. The resulting attended summary vector from this phase is the persona representation $z$, defined as:

\small
\begin{equation}
\begin{array}{c}
z=\gamma_D^s D^s+ \gamma_E^s E^s+ \gamma_O^s O^s \\
\gamma_D^s+\gamma_E^s+\gamma_O^s =1
\end{array}
\label{summary}
\end{equation}
\normalsize
where $\gamma_D^s, \gamma_E^s, \gamma_O^s$ are learnable weight parameters. $D^s,E^s,O^s$ refers to summary vectors of the $N_{diag}$ character's lines, contextual information, and other characters' lines, respectively. In Section \ref{expt}, we experiment with models that have $\gamma_E^s$ and $\gamma_O^s$ set to 0 to understand how the contextual information and other characters' lines contribute to the overall performance.

\subsubsection{Encoder}
\label{encoder}
Given an input sequence $(x_1, x_2, . . . , x_T )$, we use a recurrent neural network to encode the sequence into hidden states $(h_1, h_2, . . . , h_T)$.  In our experiments, we use a gated recurrent network (GRU)
\cite{chung2014empirical} over LSTMs \cite{hochreiter1997long} because the latter is more computationally expensive. We use bidirectional GRUs and concatenate our forward and backwards hidden states to get $\overleftrightarrow{h_t}$ for  $t=1,...,T$.

\subsubsection{Attention}
\label{attention}
We define an attention mechanism $Attn$ that computes $s$ from the resultant hidden states $\overleftrightarrow{h_t}$ of a GRU by learning to generate weights $\alpha_t$. This can be interpreted as the relative importance given to a hidden state $h_t$ to form an overall summary vector for the sequence. Formally, we define it as:

\setlength{\abovedisplayskip}{0cm}
\setlength{\belowdisplayskip}{0.5cm}
\small
\begin{gather}
a_t=f_{attn}(h_t)  \\
\alpha_t=softmax(a_t)\\
s=\sum_{t=1}^{T}{\alpha_t h_t}  
\label{softmax}
\end{gather}
\normalsize
where $f_{attn}$ is a two layer fully connected network in which the first layer projects $h_t \in {\rm I\!R}^{d_h}$  to an attention hidden space $g_t \in {\rm I\!R}^{d_a}$, and the second layer produces a relevance score  for every hidden state at timestep $t$.


\subsection{Memory Modules}
Our model consists of a read-only `Knowledge-Store' memory, and we also test a recent read-write memory. External memories have been shown to help on natural language processing tasks \cite{sukhbaatar2015end,kumar2016ask,kaiser2017aurko}, and we find similar improvements in learning capability.

\subsubsection{Knowledge-Store Memory}
The main motivation behind the Knowledge-Store memory module is to incorporate prior domain knowledge. In our work, this knowledge refers to the trope descriptions described in Section \ref{trope_desc}.

Related works have initialized their memory networks with positional encoding using word embeddings \cite{sukhbaatar2015end, kumar2016ask, miller2016key}. To incorporate the descriptions, we represent them with skip thought vectors \cite{kiros2015skip} and use them to initialize the memory keys $K_M \in {\rm I\!R}^{N_P \times d_K}$, where $N_P$ is the number of tropes, and $d_K$ is set to the size of embedded trope description $R^D$, i.e. $d_K=||R^D||$. 

The values in the memory represent learnable embeddings of corresponding trope categories $V_M \in {\rm I\!R}^{N_P \times d_V}$, where $d_V$ is the size of the trope category embeddings. The network learns to use the persona representation $z$ from the encoder phase to find relevant matches in the memory. This corresponds to calculating similarities between $z$ and the keys $K_M$. Formally, this is calculated as:

\setlength{\abovedisplayskip}{0cm}
\small
\begin{gather}
z_M=f_z(z)  \\
p_{M}^{(i)}=softmax(z_M \cdot K_M[i]) \\
 \forall i \in \{1,..,N_P\} \nonumber
\end{gather}
\normalsize
where $f_z: {\rm I\!R}^{d_h} \mapsto {\rm I\!R}^{d_K} $ is a fully-connected layer that projects the persona representation in the space of memory keys $K_M$. 
Based on the match probabilities $p_M^{(i)}$, the values $V_M$ are weighted and cumulatively added to the original persona representation as:

\setlength{\abovedisplayskip}{0cm}
\small
\begin{equation}
r_{out}=\sum_{i=1}^{N_P}{p_M^{(i)}\cdot V_M[i]} 
\end{equation}
\normalsize

We iteratively combine our mapped persona representation $z_M$ with information from the memory $r_{out}$. The above process is repeated $n_{hop}$ times. The memory mapped persona representation $z_M$ is updated as follows:

\small
\begin{equation}
z_M^{{hop}}=f_r(z_M^{{hop}-1}) + r_{out} 
\end{equation}
\normalsize
where $z_M^0=z_M$, and $f_r :  {\rm I\!R}^{d_V} \mapsto {\rm I\!R}^{d_K}$ is a fully-connected layer. Finally, we transform the resulting $z_M^{n_{hop}}$ using another fully-connected layer, $f_{out} \in  {\rm I\!R}^{d_K} \mapsto {\rm I\!R}^{d_h}$, via:

\small
\begin{equation}
\hat{z}_M=f_{out}(z_M^{n_{hop}}) 
\end{equation}
\normalsize

\subsubsection{Read-Write Memory}
We also tested a Read-Write Memory following Kaiser et. al \cite{kaiser2017aurko}, which was originally designed to remember rare events. In our case, these `rare' events might be key dialogue snippets that are particularly indicative of latent persona. It consists of keys, which are activations of a specific layer of model, i.e. the persona representation $z$, and values, which are the ground-truth labels, i.e. the trope categories. Over time, it is able to facilitate predictions based on past data with similar activations stored in the memory. For every new example, the network writes to memory for future look up. A memory with memory size $N_M$ is defined as:

\small
\begin{equation}
M=(K_{N_M \times d_H}, V_{N_M} , A_{N_M} ) 
\end{equation}
\normalsize

\paragraph{Memory Read} We use the persona embedding $z$ as a query to the memory. We calculate the cosine similarities between $z$ and the keys in $M$, take the softmax on the top-k neighbors, and compute a weighted embedding $\hat{z}_M$ using those scores.

\paragraph{Memory Write} 
We update the memory in a similar fashion to the original work by \cite{kaiser2017aurko}, which takes into account the maximum age of items as stored in $A_{N_M}$.

\section{Objective Losses}
To train our model, we utilize the different objective losses described below.

\subsection{Classification Loss}
We calculate the probability of a character belonging to a particular trope category $P$ through Equation \ref{eqn_class_softmax}, where $f_{P} :  {\rm I\!R}^{d_h} \mapsto {\rm I\!R}^{N_P}$ is a fully-connected layer, and $z$ is the persona representation produced by the multi-level attentive encoders described in Equation \ref{summary}. We then optimize the categorical cross-entropy loss between the predicted and true tropes as in Equation \ref{cat_ce}, where $N_P$ is the total number of tropes, $q_j$  is the predicted distribution that the input character fulls under trope $j$, and $p_{j} \in \{0,1\}$ denotes the ground-truth of whether the input snippets come from characters from the $j^{th}$ trope.

\setlength{\abovedisplayskip}{0cm}
\small
\begin{align}
q = softmax(f_{P}(z)) \label{eqn_class_softmax} \\
J_{CE}= \sum_{j=1}^{N_P}{-p_{j}\log(q_{j})} \label{cat_ce}
\end{align}
\normalsize

\subsection{Trope Description Triplet Loss}
In addition to using trope descriptions to initialize the Knowledge-Store Memory, we also test learning from the trope descriptions through a triplet loss \cite{hoffer2015deep}. We again use the skip thought vectors to represent the descriptions. Specifically, we want to maximize the similarity of representations obtained from dialogue snippets with their corresponding description, and minimize their similarity with negative examples. We implement this as:

\setlength{\abovedisplayskip}{0cm}
\small
\begin{gather}
R^P = f_{D}(z)
\label{f_D} \\
J_{T} = max(0,s(R^P, R^D_n) - s(R^P, R^D_p) + \alpha_{T}) \label{TL}
\end{gather}
\normalsize
where $f_{D} :  {\rm I\!R}^{d_h} \mapsto {\rm I\!R}^{||R^D||}$ is a fully-connected layer. The triplet ranking loss is then Equation \ref{TL}, where $\alpha_{T}$ is a learnable margin parameter and $s(\cdot, \cdot)$ denotes the similarity between trope embeddings ($R^P$), positive ($R^D_p$) and negative ($R^D_n$) trope descriptions.

\subsubsection*{Trope Description Triplet Loss with Memory Module}
If a memory module is used, we compute a new triplet loss in place of the one described in Equation \ref{TL}. Models that use a memory module should learn a representation $\hat{z}_M$, based on either the prior knowledge stored in the memory (as in Knowledge-Store memory) or the top-$k$ key matches (as in Read-Write memory), that is similar to the representation of the trope descriptions.

This is achieved by replacing the persona embedding $z$ in Equation \ref{f_D} with the memory output $\hat{z}_M$ as shown in Equation \ref{f_d_m}, where $f_{D_M}:  {\rm I\!R}^{d_h} \mapsto {\rm I\!R}^{||R^D||}$ is a fully-connected layer. To compute the new loss, we combine the representations obtained from Equations \ref{f_D} and \ref{f_d_m} through a learnable parameter $\gamma$ that determines the importance of each representation. Finally, we utilize this combined  representation $\hat{R}^P$ to calculate the loss as shown in Equation \ref{jmt}.

\setlength{\abovedisplayskip}{0cm}
\small
\begin{gather}
R^P_M=f_{D_M}(\hat{z}_M) \label{f_d_m}\\
\hat{R}^P=\gamma R^P+(1-\gamma) R_M^P \label{hatR}\\
J_{MT}=max(0, s(\hat{R}^P, R^D_n) - s(\hat{R}^P, R^D_p) + \alpha_{MT}) \label{jmt}
\end{gather}
\normalsize

\subsection{Read-Write Memory Losses}
When the Read-Write memory is used, we use two extra loss functions. The first is a Memory Ranking Loss $J_{MR}$ as done in \cite{kaiser2017aurko}, which learns based on whether a query with the persona embedding $z$ returns nearest neighbors with the correct trope.  The second is a Memory Classification Loss $J_{MCE}$ that uses the values returned by the memory to predict the trope. The full details for both are found in Supplementary Section \ref{sec_rw_mem_losses}.

\subsection{Overall Loss}
We combine the above losses through:
\setlength{\abovedisplayskip}{0cm}
\small
\begin{equation}
\begin{array}{c}
J=\beta_{CE}\cdot J_{CE}\\
+ \;\beta_{T}\cdot \hat{J}_{T}\\
+ \; \beta_{MR}\cdot J_{MR}+ \; \beta_{MCE}\cdot J_{MCE}  \\
\\
  \hat{J}_T=\begin{cases}
    J_{MT} & \text{if memory module is used }\\
    J_{T} & \text{otherwise}.
  \end{cases}
\end{array} 
\label{main_loss}
\end{equation}
\normalsize
where $\beta =[\beta_{CE}, \beta_{MCE}, ,\beta_{T},  \beta_{MR} ]$ are learnable weights such that $\sum_{i}\beta_i=1$. Depending on which variant of the model is being used, the list $\beta$ is modified to contain only relevant losses. For example, when the Knowledge-Store memory is used, we set $\beta_{MR} =\beta_{MCE}=0$ and  $\beta$ is modified to $\beta=[\beta_{CE}, \beta_{T}]$. We discuss different variants of our model in the next section.

\begin{table*}
\small
\begin{center}
\npdecimalsign{.}
\nprounddigits{3}
\begin{tabular}{|c|n{1}{5}|n{1}{5}|n{1}{5}|n{1}{5}|}
\hline
\bf{Model} & \bf{Accuracy} & \bf{Precision} & \bf{Recall} & \bf{F1} \\ \hline
	    baseline\_char & 0.286057692308 & 0.538461538462 & 0.286153846154 & 0.339230769231 \\
    baseline\_3 & 0.287259615385 & 0.552307692308 & 0.288461538462 & 0.349230769231 \\
    attn\_char & {\npboldmath}0.629807692308 & 0.586153846154 & 0.627692307692 &      0.6       \\
    attn\_3  & 0.629807692308 & {\npboldmath}0.59 & {\npboldmath}0.63 & {\npboldmath}0.603076923077 \\
    \hline 
    attn\_3\_tropetrip & 0.615384615385 & 0.566153846154 & 0.614615384615 & 0.583076923077 \\
    attn\_3\_tropetrip-500 & {\npboldmath}0.644230769231 & {\npboldmath}0.600769230769 & {\npboldmath}0.642307692308 & {\npboldmath}0.614615384615 \\
    \hline
    attn\_3\_ks-mem & {\npboldmath}0.677884615385 & {\npboldmath}0.647692307692 & {\npboldmath}0.676153846154 & {\npboldmath}0.656923076923 \\
    attn\_3\_rw-mem  & 0.634615384615 & 0.598461538462 & 0.634615384615 & 0.610769230769 \\
    \hline
    attn\_3\_tropetrip\_ks-mem & {\npboldmath}0.663461538462 & {\npboldmath}0.628461538462 & {\npboldmath}0.662307692308 & {\npboldmath}0.639230769231 \\
    attn\_3\_tropetrip-500\_ks-mem & 0.653846153846 & 0.617692307692 & 0.652307692308 & 0.629230769231 \\
    attn\_3\_tropetrip\_rw-mem & 0.649038461538 & 0.601538461538 & 0.647692307692 & 0.616923076923 \\
    attn\_3\_tropetrip-500\_rw-mem  & 0.644230769231 & 0.607692307692 & 0.643846153846 & 0.619230769231 \\
    \hline
    attn\_3\_tropetrip-500\_ks-mem\_ndialog16 & 0.740384615385 & 0.706923076923 & 0.740769230769 & 0.717692307692 \\
    \textbf{attn\_3\_tropetrip-500\_ks-mem\_ndialog32} & {\npboldmath}0.75 & {\npboldmath}0.75  & {\npboldmath}0.75 & {\npboldmath}0.75 \\
    attn\_3\_tropetrip\_ks-mem\_ndialog16 & 0.740384615385 & 0.721538461538 & 0.740769230769 & 0.727692307692 \\
    attn\_3\_tropetrip\_ks-mem\_ndialog32 & 0.730769230769 & 0.711538461538 & 0.730769230769 & 0.718461538462 \\
\hline
\end{tabular}
\npnoround
\end{center}
\caption{\label{results} Experimental results. Details and analysis are given in Section \ref{ablation_sec}. The best performing results in each block are bolded. The first block examines the baseline model vs. the attention model, as well as use of different inputs. The second block uses the triplet loss, and the third block uses our memory modules. The fourth block combines the triplet loss and memory module, which the fifth block extends to larger $N_{diag}$.}
\end{table*}

\section{Experiments}
\label{expt}

We experimented with combinations of our various modules and losses. The experimental results and ablation studies are described in the following sections, and the experimental details are described in Supplementary Section \ref{sec_exp_details}. The different model permutation names in Table \ref{results}, e.g. ``attn\_3\_tropetrip\_ks-mem\_ndialog16'', are defined as follows:
\begin{itemize}
  \setlength{\itemsep}{1pt}
  \setlength{\parskip}{0pt}
  \setlength{\parsep}{0pt}
  \item \textit{baseline vs attn}: The `baseline' model uses only one dialogue snippet $S$ to predict the trope, i.e. $N_{diag}=1$. Hence, the inter-snippet encoder is not used.  The `attn' model operates on $N_{diag}$ dialogue snippets using the inter-snippet encoder to assign an attention score for each snippet $S_i$.
  \item \textit{char vs. 3}: To measure the importance of context and other characters' lines, we have two variants -- `char' uses only the character's lines, while `3' uses the character's lines, other character's lines, and all context lines. Formally, in `char' mode, we set $\gamma_E^s$ and  $\gamma_O^s$ to 0 in Equation \ref{summary}. In `attn' mode, $(\gamma_E^s,\gamma_O^s, \gamma_D^s)$ are learned by the model.
 
  \item \textit{tropetrip}: The presence of 'tropetrip' indicates that the triplet loss on the trope descriptions was used. If `-500' is appended to `tropetrip', then the 4800-dimensional skip embeddings representing the descriptions in Equations \ref{f_d_m} and \ref{jmt} are projected to 500 dimensions using a fully connected layer.
  
  \item \textit{ks-mem vs. rw-mem}: `ks-mem' refers to the Knowledge-Store memory, and `rw-mem' refers to the Read-Write memory.
  
  \item \textit{ndialog}: The number of dialogue snippets $N_{diag}$ used as input for the attention models. Any attention model without the explicit $N_{diag}$ listed uses $N_{diag}=8$.
\end{itemize}

\subsection{Ablation Results}\label{ablation_sec}
\textbf{Baseline vs. Attention Model.}
The attention model shows a large improvement over the baseline models. This matches our intuition that not every quote is strongly indicative of character trope. Some may be largely expository or `chit-chat' pieces of dialogue. Example attention scores are shown in Section \ref{attention_scores_sec}.

Though our experiments showed marginal improvement between using the `char' data and the `3' data, we found that using all 3 inputs had greater performance for models with the triplet loss and read-only memory. This is likely because the others' lines and context capture more of the social dynamics and situations that are described in the trope descriptions. Subsequent results are shown only for the `attn\_3' models.

\begin{figure}
\centering
\includegraphics[width=1.0\columnwidth]{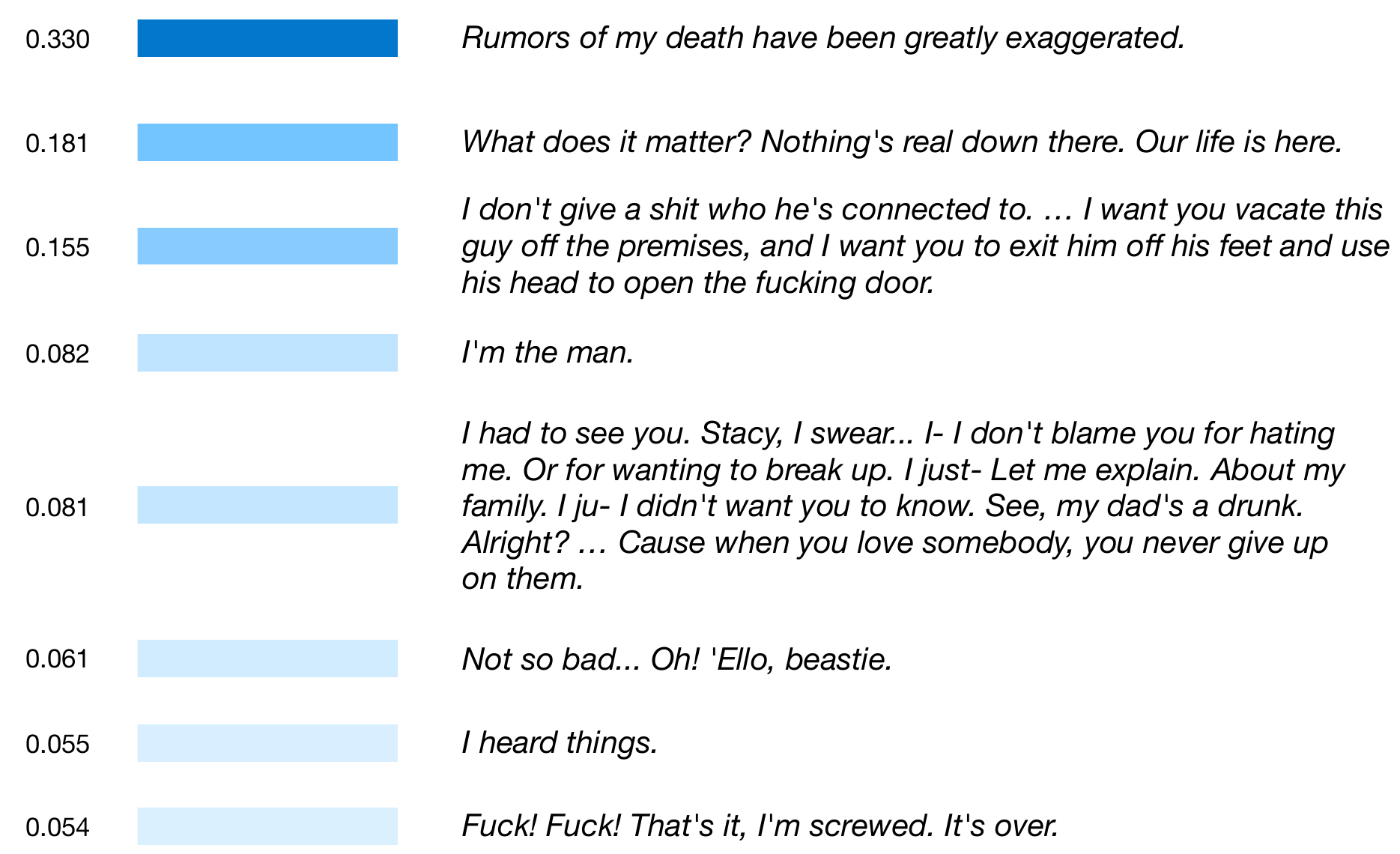}
\caption{\label{attn_scores_fig}Attention scores for a batch of dialogues for the ``byronic hero'' trope \footnotemark.}
\end{figure}
\footnotetext{TVtropes.org defines a byronic hero as ``Sometimes an Anti-Hero, others an Anti-Villain, or even Just a Villain, Byronic heroes are charismatic characters with strong passions and ideals, but who are nonetheless deeply flawed individuals....''}

\textbf{Trope Description Triplet Loss.}
Adding the trope description loss alone provided relatively small gains in performance, though we see greater gains when combined with memory. While both use the descriptions, perhaps the Knowledge Store memory matches an embedding against all the tropes, whereas the trope triplet loss is only provided information from one positive and one negative example.

\textbf{Memory Modules.}
The Knowledge-Store memory in particular was helpful. Initialized with the trope descriptions, this memory can `sharpen' queries toward one of the tropes. The Read-Write memory had smaller gains in performance. It may be that more data is required to take advantage of the write capabilities.

\textbf{Combined Trope Description Triplet Loss and Memory Modules.}
Using the triplet loss with memory modules led to greater performance when compared to the $attn\_3$ model, but the performance sits around the use of either triplet only or memory only. However, when we increase the $N_{diag}$ to 16 or 32, we find a jump in performance. This is likely the case because the model has both increased learning capacity and a larger sample of data at every batch, which means at least some of the $N_{diag}$ quotes should be informative about the trope.


 

\subsection{Attention Scores}\label{attention_scores_sec}
Because the inter-snippet encoder provides such a large gain in performance compared to the baseline model, we provide an example illustrating the weights placed on a batch of $N_{diag}$ snippets. Figure \ref{attn_scores_fig} shows the attention scores for the character's lines in the ``byronic hero'' trope. Matching what we might expect for an antihero personality, we find the top weighted line to be full of confidence and heroic bluster, while the middle lines hint at the characters' personal turmoil. We also find the lowly weighted sixth and seventh lines to be largely uninformative (e.g. ``I heard things.''), and the last line to be perhaps too pessimistic and negative for a hero, even a byronic one.

\subsection{Purity scores of character clusters}
Finally, we measure our ability to recover the trope `clusters' (with one trope being a cluster of its characters) with our embeddings through the purity score used in \cite{bamman2014learning}. Equation \ref{eqn_purity} measures the amount of overlap between two clusterings, where $N$ is the total number of characters, $g_i$ is the $i$-ith ground truth cluster, and $c_j$ is the $j$-th predicted cluster.

\setlength{\abovedisplayskip}{0cm}
\small
\begin{equation}
\text{Purity} = \frac{1}{N}\sum_i \text{max}_j |g_i \cap c_j|
\label{eqn_purity}
\end{equation}
\normalsize

We use a simple agglomerative clustering method on our embeddings with a parameter $k$ for the number of clusters. The methods in \cite{bamman2014learning} contain a similar hyper-parameter for the number of persona clusters. We note that the metrics are not completely comparable because not every character in the original dataset was found on IMDB. The results are shown in Table \ref{tab_purity}. It might be expected that our model perform better because we use the character tropes themselves as training data. However, dialogue may be noisier than the movie summary data; their better performing Persona Regression (PR) model also uses useful metadata features such as the movie genre and character gender. We simply note that our scores are comparable or higher.

\begin{table}[h!]
\small
\begin{center}
\begin{tabular}{|c|c|c|c|}
\hline $k$ & PR & DP & AMN \\ \hline
25 & 42.9 & 39.63 & \textbf{48.4} \\
50 & 36.5 & 31.0 &  \textbf{48.1} \\
100 & 30.3 & 24.4 & \textbf{45.2} \\
\hline
\end{tabular}
\end{center}
\caption{Cluster purity scores. $k$ is the number of clusters, PR and DP are the Persona Regression and Dirichlet Persona models from \cite{bamman2014learning}, and AMN is our attention memory network.}
\label{tab_purity}
\normalsize
\end{table}

\section{Application: Narrative Analysis}
We collected IMDB quotes for the top 250 movies on IMDB. For every character, we calculated a character embedding by taking the average embedding produced by passing all the dialogues through our model. We then calculated movie embeddings by taking the weighted sum of all the character embeddings in the movie, with the weight as the percentage of quotes they had in the movie. By computing distances between pairs of character or movie embeddings, we could potentially unearth notable similarities. We note some of the interesting clusters below.

\subsection{Clustering Characters}
\begin{itemize}
  \setlength{\itemsep}{1pt}
  \setlength{\parskip}{0pt}
  \setlength{\parsep}{0pt}
  \item Grumpy old men: Carl Fredricksen (Up); Walk Kowalski (Gran Torino)
  \item Shady tricksters, crooks, well versed in deceit: Ugarte (Casablanca); Eames (Inception)
  \item Intrepid heroes, adventurers: Indiana Jones (Indiana Jones and the Last Crusade); Nemo (Finding Nemo); Murph (Interstellar)
\end{itemize}

\subsection{Clustering Movies}
\begin{itemize}
  \setlength{\itemsep}{1pt}
  \setlength{\parskip}{0pt}
  \setlength{\parsep}{0pt}
  \item Epics, historical tales: Amadeus, Ben-Hur
  \item Tortured individuals, dark, violent: Donnie Darko, Taxi Driver, Inception, The Prestige
  \item Gangsters, excess: Scarface, Goodfellas, Reservoir Dogs, The Departed, Wolf of Wall Street
\end{itemize}

\section{Conclusion}
We used the character trope classification task as a test bed for learning personas from dialogue. Our experiments demonstrate that the use of a multi-level attention mechanism greatly outperforms a baseline GRU model. We were also able to leverage prior knowledge in the form of textual descriptions of the trope. In particular, using these descriptions to initialize our Knowledge-Store memory helped improved performance. Because we use short text and can leverage domain knowledge, we believe future work could use our models for applications such as personalized dialogue systems.

\bibliography{ours}
\bibliographystyle{acl_natbib_nourl}

\clearpage
\appendix

\section{Read-Write Memory Losses}
\label{sec_rw_mem_losses}

\subsection{Memory Ranking Loss $J_{MR}$}
We want the model to learn to make efficient matches with the memory keys in order to facilitate look up on past data. To do this, we find a positive and negative neighbor after computing the $k$ nearest neighbors $(n_1, . . . , n_k)$ by finding the smallest index $p$ such that $V[p] = P$ and $n$ such that $V[n] \neq P$ respectively. We define the memory ranking loss as:

\setlength{\abovedisplayskip}{0cm}
\setlength{\belowdisplayskip}{0.5cm}
\small
\begin{equation}
J_{MR}=max(0,s(z, K[n]) - s(z, K[p]) + \alpha_{MR})  
\end{equation}
\normalsize
where $\alpha_{MR}$ is a learnable margin parameter and $s(\cdot, \cdot)$ denotes the similarity between persona embeddings ($z$), key representations of positive ($K[p]$) and negative ($K[n]$) neighbors.
The above equation is consistent with the memory loss defined in the original work
\cite{kaiser2017aurko}.

\subsection{Memory Classification Loss $J_{MCE}$} The Read-Write memory returns $\hat{z}_M$ and values $v_M=V[n_i] \; \forall i\in \{n_1,..,n_k\}$. 
The probability of the given input dialogues belonging to a particular persona category $P$ is computed using values $v_M$ returned from the memory via:

\small
\begin{equation}
q^M=softmax(f_{{P}_M}(v_M)) 
\label{qm}
\end{equation}
\normalsize
where $f_{{P}_M} :  {\rm I\!R}^{k} \mapsto {\rm I\!R}^{N_P}$ is a fully-connected layer. We replace the $q_j$ with $q_j^M$ in Equation \ref{cat_ce} and calculate the categorical cross entropy to get $J_{MCE}$.


\section{Experimental details}
\label{sec_exp_details}

The vocabulary size was set to 20000. We used a GRU hidden size of 200, a word embedding size of 300, and the word embedding lookup was initialized with GLoVe \cite{pennington2014glove}. For the Read-Write memory module, we used k=8 when calculating the nearest neighbors and a memory size of 150. Our models were trained using Adam \cite{kingma2014adam}.

\end{document}